\pdfoutput=1

\documentclass[11pt]{article}

\usepackage[]{ACL2023}

\usepackage{times}
\usepackage{latexsym}

\newcommand\cincludegraphics[2][]{\raisebox{-\height}{\includegraphics[#1]{#2}}}

\usepackage[T1]{fontenc}

\usepackage[utf8]{inputenc}

\usepackage{microtype}

\usepackage{inconsolata}

\usepackage{listings}
\usepackage{microtype}
\usepackage{verbatim}
\usepackage{booktabs}
\usepackage{cellspace}
\usepackage{arydshln}
\usepackage{amssymb}
\usepackage{amsmath}
\usepackage{graphicx}
\usepackage{bm}

\usepackage{times}
\usepackage{microtype}
\usepackage{epsfig}
\usepackage{caption}
\usepackage{float}
\usepackage{placeins}
\usepackage{color, colortbl}
\usepackage{stfloats}
\usepackage{enumitem}
\usepackage{tabularx}
\usepackage{xstring}
\usepackage{multirow}
\usepackage{xspace}
\usepackage{url}
\usepackage{subcaption}
\usepackage{xcolor}

\title{Toward Expanding the Scope of Radiology Report Summarization to Multiple Anatomies and Modalities}

\author{
    Zhihong Chen$^{2,3*}$, \hspace{0.2cm} 
    Maya Varma$^{1*}$, \hspace{0.2cm}
    Xiang Wan$^{2,3}$, \hspace{0.2cm} \\
    \textbf{Curtis P. Langlotz}$^{1}$,\hspace{0.2cm} \textbf{Jean-Benoit Delbrouck}$^{1*}$\\
    $^{1}$Stanford University\\
    $^{2}$The Chinese University of Hong Kong, Shenzhen\\
    $^{3}$Shenzhen Research Institute of Big Data\\
    \texttt{zhihongchen@link.cuhk.edu.cn}  \hspace{0.2cm} \texttt{wanxiang@sribd.cn}  \\ \texttt{\{mvarma2,langlotz,jbdel\}@stanford.edu}     
}

\begin{document}
\maketitle
\renewcommand{\thefootnote}{\fnsymbol{footnote}}
\footnotetext[1]{Equal Contribution.}
\renewcommand{\thefootnote}{\arabic{footnote}}

\begin{abstract}
Radiology report summarization (RRS) is a growing area of research.
Given the Findings section of a radiology report, the goal is to generate a summary (called an Impression section) that highlights the key observations and conclusions of the radiology study.
However, RRS currently faces essential limitations.
First, many prior studies conduct experiments on private datasets, preventing the reproduction of results and fair comparisons across different systems and solutions.
Second, most prior approaches are evaluated solely on chest X-rays.
To address these limitations, we propose a dataset (MIMIC-RRS) involving three new modalities and seven new anatomies based on the MIMIC-III and MIMIC-CXR datasets.
We then conduct extensive experiments to evaluate the performance of models both within and across modality-anatomy pairs in MIMIC-RRS. In addition, we evaluate their clinical efficacy via RadGraph, a factual correctness metric.
\end{abstract}

\section{Introduction}
A \textit{radiology report} is a document that provides information about the results of a radiology study. It usually includes a Findings section with key observations from the study and an Impression section with the radiologist's overall conclusions. The latter is the most critical part of the report and is typically based on both the findings and the patient's condition. It can be helpful to automate the process of generating the impression section because it can be time-consuming and prone to errors when done manually~\cite{bhargavan2009workload,alexander2022mandating}. Recently, substantial progress has been made towards research on automated radiology report summarization (RRS)~\citep{zhang2020optimizing,ben-abacha-etal-2021-overview,hu2022graph}.

However, the field of RRS faces several key limitations.
First, the experimental results of many prior studies~\cite{zhang2018learning,zhang2020optimizing} are reported on private datasets, making it difficult to replicate results or compare approaches. Second, existing studies are mainly limited to a single modality (\textit{i.e.}, X-ray) and a single anatomy (\textit{i.e.}, chest)~\cite{zhang2020optimizing,ben-abacha-etal-2021-overview,hu-etal-2021-word}. In some cases, researchers omit to disclose the modality and anatomy of the radiology reports used for their experiments~\citep{karn2022differentiable}.
Finally, recent models~\citep{karn2022differentiable,hu2022graph} present an increased complexity in architecture that offers only marginal improvements on the existing evaluation metrics for summarization. This further makes the replication of studies more difficult.

To address the aforementioned limitations, we construct a brand-new open-source dataset (named MIMIC-RRS) for radiology report summarization involving three modalities (X-ray, MRI, and CT) and seven anatomies (chest, head, neck, sinus, spine, abdomen, and pelvis). MIMIC-RRS is based on the MIMIC-CXR~\cite{johnson2019mimic} and MIMIC-III~\citep{johnson2016mimic} datasets and introduces data from 12 new modality-anatomy pairs.
As a result, we introduce a new setting for evaluating the generalization capabilities of RRS models across different modalities and anatomies.

In addition, we benchmark various pre-trained language models on MIMIC-RRS.
Through extensive experiments within and across modality-anatomy pairs, we show that adopting an appropriate pre-trained model can achieve promising results comparable to previous studies. We also introduce a metric to evaluate factual correctness of generated summaries for any modality-anatomy pair.

\section{Dataset Construction}
In this section, we present the new MIMIC-RRS dataset designed for radiology report summarization across multiple modalities and anatomies. Comparisons with existing datasets are shown in Table~\ref{table:existing-datasets}. We detail the collection process and the dataset statistics in the following subsections.

\begin{table*}[t]
\centering
\begin{tabular}{@{}lrrcr@{}}
\toprule
Dataset                          & Anatomy  & Modality & Language & Number  \\ \midrule
\citet{zhang2018learning}        & Multiple & Multiple & English  & 87,127  \\
\citet{zhang2020optimizing}      & Multiple & Multiple & English  & 130,850 \\
RIH~\cite{zhang2020optimizing}   & Multiple & Multiple & English  & 139,654 \\
OpenI~\cite{demner2016preparing} & Chest    & X-ray    & English  & 3,268   \\
MIMIC-CXR~\cite{johnson2019mimic}                & Chest    & X-ray    & English  & 128,003 \\
PadChest~\cite{bustos2020padchest}& Chest    & X-ray    & Spanish  & 206,222 \\ \midrule
MIMIC-RRS (ours)                 & Multiple & Multiple & English  & 207,782 \\ \bottomrule
\end{tabular}
\caption{Comparisons with existing datasets for radiology report summarization.}
\label{table:existing-datasets}
\end{table*}

\subsection{Data Collection}
\paragraph{MIMIC-III} One of our main contributions is to generate RRS data from MIMIC-II involving distinct combinations of modalities (\textit{i.e.}, medical imaging techniques) and anatomies (\textit{i.e.}, body parts). To this end, we first select five of the most frequently-occurring modality-anatomy pairs in the pool of MIMIC-III reports: ``CT Head'', ``CT Spine'', ``CT Chest'', ``CT Abdomen-Pelvis'' and ``MR Head''. Note that we discard chest X-rays as they are included in the MIMIC-CXR dataset. In addition, we pick six modality-anatomy pairs that occur infrequently in MIMIC-III to serve as out-of-domain (OOD) test sets: ``CT Neck'', ``CT Sinus'', ``MR Pelvis'',  ``MR Neck'',  ``MR Abdomen'', ``MR Spine''. This set of pairs represents two types of OOD cases: (1) the modality has not been seen during training (one could train on CT neck and test on MR Neck), and (2) the anatomy has not been seen during training (for example, CT Sinus is the only ``sinus'' dataset).

For each report, we extract the findings and impression section. However, the findings section is not always clearly labeled as ``findings". With the help of a board-certified radiologist, we identify alternate section headers that reference findings for each modality-anatomy pair. As an example, for CT head, findings may be referenced in reports with the section headings ``\textit{non-contrast head ct}", ``\textit{ct head}", ``\textit{ct head without contrast}", ``\textit{ct head without iv contrast}", ``\textit{head ct}", ``\textit{head ct without iv contrast}", or ``\textit{cta head}". We identify 537 candidate section headers that reference findings across our dataset. We also discarded reports where multiple studies are pooled in the same radiology report, leading to multiple intricate observations in the impression section\footnote{We release our candidate section headers as well as code to recreate the dataset from scratch (Appendix~\ref{app:code}).}. Our resulting dataset consists of 79,779 selected reports across 11 modality-anatomy pairs.

\paragraph{MIMIC-CXR} MIMIC-CXR studies are chest X-ray examinations. We follow preprocessing steps reported in previous work~\cite{delbrouck-etal-2022-vilmedic}, and we only include reports with both a Findings and an Impression section. This yields 128,003 reports.

\subsection{Data statistics}
\begin{figure}[t]
  \centering\includegraphics[width=1.0\linewidth]{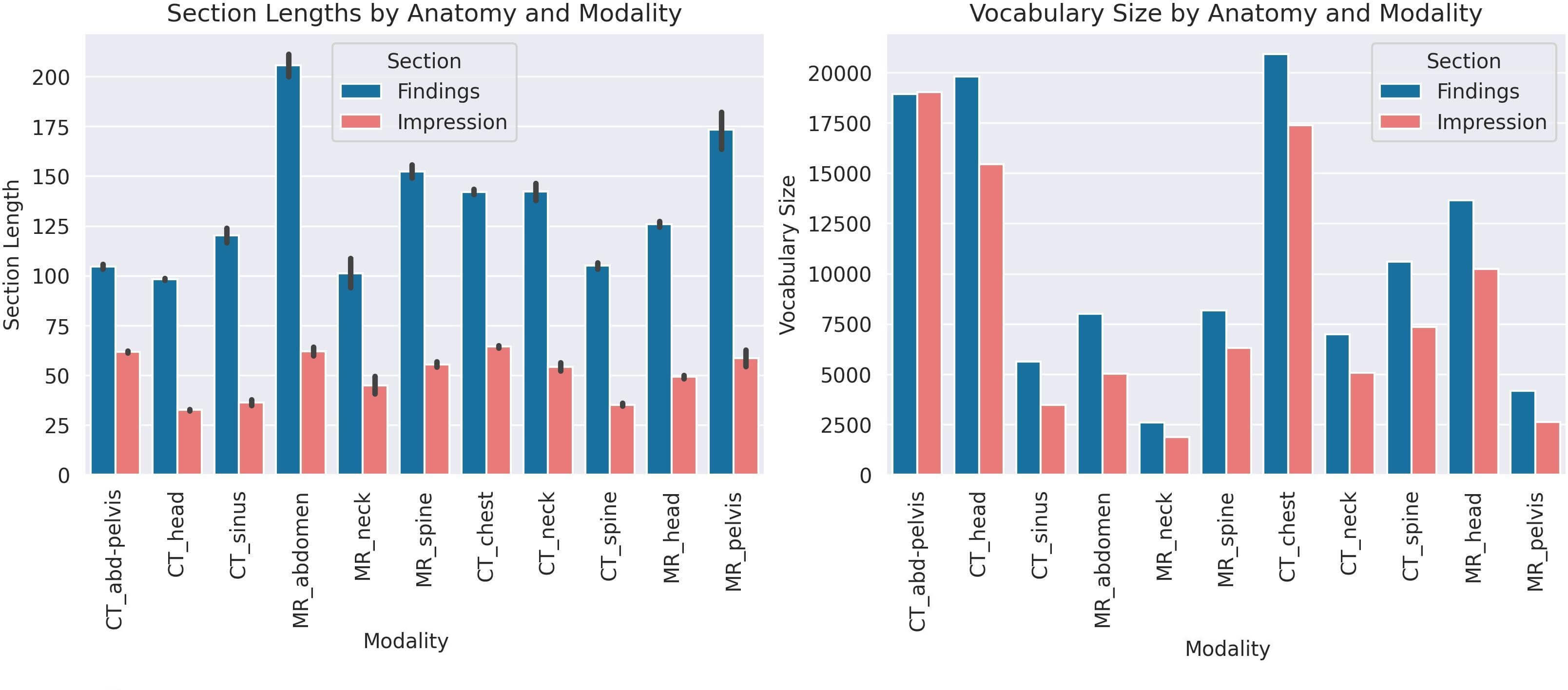}
  \caption{Section length and vocabulary size for reports from each modality-anatomy pair.}
  \label{fig:stats}
\end{figure}

\begin{table}[t]
\centering
\resizebox{0.8\linewidth}{!}{
\begin{tabular}{@{}cccc@{}}
\toprule
CT Abd-pelv & CT Chest  & CT Head    \\ \midrule
15,989      & 12,786    & 31,402     \\ \midrule
CT Spine    & MR Head   & CT Neck    \\ \midrule
5,517       & 7,313     & 1,140       \\ \midrule
CT Sinus    & MR Spine  & MR Abdomen \\\midrule
1,267       & 2,821     & 1,061       \\ \midrule
MR Neck     & MR Pelvis & X-ray Chest   \\\midrule
230         & 253       & 128,003 \\ \bottomrule
\end{tabular}}
\caption{Dataset statistics for MIMIC-RRS. We report the number of radiology reports from each modality-anatomy pair.}
\label{tab:numsamples}
\end{table}
In total, there are 207,782 samples in the MIMIC-RRS dataset. The number of examples for each modality and anatomy is provided in Table~\ref{tab:numsamples}. 
To further analyze this dataset, we report  in Figure~\ref{fig:stats} the text lengths and vocabulary sizes associated with reports from each modality-anatomy pair.
We find that for all modality-anatomy pairs, the findings section is significantly longer than the impression section (up to +315\% for MR abdomen).
Additionally, the findings sections of chest X-ray reports, which average only 49 words, are much shorter than reports from other modality-anatomy pairs. In contrast, MR Abdomen and MR Pelvis reports including findings sections that average 205 and 174 words, respectively. We see that CT Chest, CT Head, and CT Abdomen-Pelvis reports have a relatively large vocabulary size (given their sample size) with 20,909, 19,813, and 18,933 words. Surprisingly, the CT Abdomen-Pelvis impressions include a larger vocabulary than the findings. On the other hand, MR pelvis and MR abdomen impressions contain 36\% and 37\% fewer words than their corresponding findings, respectively. 

We assign reports from the following modality-anatomy pairs to training, validation, and test sets due to their large sample sizes: ``CT abdomen/pelvis'', ``CT Chest'', ``CT Neck'', ``CT Spine'', ``CT Head'', ``MR Head'', and ``X-ray Chest''. The remaining reports (\textit{i.e.}, ``MR Pelvis'', ``MR Spine'', ``MR Neck'', ``MR Abdomen'', and ``CT Sinus'') are used for OOD test sets\footnote{We release data splits publicly so that future work can fairly compare new results.}. 

\section{Algorithmic Analysis}
In this section, we conduct experiments to analyze the performance of different models on MIMIC-RRS. We provide three categories of analyses: in-modality-anatomy, cross-modality-anatomy, and clinical efficacy.

\subsection{In-modality-anatomy}
\begin{table*}[t]
\footnotesize
\centering
\setlength{\tabcolsep}{0.8mm}{
\resizebox{0.99\linewidth}{!}{
\begin{tabular}{@{}lccccccccccccccccccccc@{}}
\toprule
\multirow{2}{*}{Models} & \multicolumn{3}{c}{MR Head}                   & \multicolumn{3}{c}{CT Spine}                  & \multicolumn{3}{c}{CT Neck}                   & \multicolumn{3}{c}{CT Head}                   & \multicolumn{3}{c}{CT Chest}                  & \multicolumn{3}{c}{CT Abd/Pel}                & \multicolumn{3}{c}{X-ray Chest}               \\
                        & R1            & R2            & RL            & R1            & R2            & RL            & R1            & R2            & RL            & R1            & R2            & RL            & R1            & R2            & RL            & R1            & R2            & RL            & R1            & R2            & RL            \\ \midrule
WGSum                   & -             & -             & -             & -             & -             & -             & -             & -             & -             & -             & -             & -             & -             & -             & -             & -             & -             & -             & 48.4          & 33.3          & 46.7          \\
AIG-CL                  & -             & -             & -             & -             & -             & -             & -             & -             & -             & -             & -             & -             & -             & -             & -             & -             & -             & -             & 51.0          & 35.2          & 46.7          \\
T5-S                    & 38.2          & 18.3          & 28.5          & 35.8          & 18.6          & 28.9          & 39.0          & 20.0          & 29.1          & 43.1          & 25.3          & 36.5          & 39.5          & 18.5          & 29.3          & 28.9          & 10.6          & 21.2          & 47.8          & 32.2          & 43.5          \\
BioBART-B               & 42.4          & 21.2          & 32.0          & 47.8          & 27.9          & 40.0          & 40.4          & 19.6          & 29.3          & 46.0          & 27.4          & 38.9          & 41.4          & 19.1          & 30.3          & 33.1          & 12.5          & 23.2          & 49.6          & 33.8          & 45.3          \\
BioBART-L               & 42.1          & 21.4          & 32.6          & 47.8          & 28.1          & 40.8          & 40.3          & 19.4          & 29.6          & 45.5          & 26.7          & 38.6          & 40.2          & 17.8          & 28.9          & 32.5          & 11.7          & 22.6          & 49.3          & 33.3          & 44.9          \\
BART-B                  & 42.0          & 21.5          & 32.1          & 49.0          & 29.7          & \textbf{41.6} & 41.4          & \textbf{20.9} & 30.2          & 46.4          & \textbf{28.1} & \textbf{39.5} & 41.6          & \textbf{19.5} & \textbf{30.6} & 33.1          & \textbf{12.9} & \textbf{23.6} & 51.0          & \textbf{34.9} & 46.4          \\
BART-L                  & \textbf{43.7} & \textbf{22.1} & \textbf{32.8} & \textbf{49.8} & \textbf{29.7} & 41.4          & \textbf{42.0} & 20.5          & \textbf{30.4} & \textbf{46.6} & 27.3          & 39.0          & \textbf{41.8} & 18.6          & 29.6          & \textbf{33.9} & 12.4          & 23.2          & \textbf{51.7} & \textbf{34.9} & \textbf{46.8} \\ \bottomrule
\end{tabular}}}
\caption{The benchmarking comparisons of different approaches, including task-specific models (\textit{i.e.}, WGSum~\cite{hu-etal-2021-word} and AIG-CL~\cite{hu2022graph}) and pre-trained language models (\textit{i.e.}, T5-S, BioBART-B, BioBART-L, BART-B, and BART-L). R1, R2, and RL denote ROUGE-1, ROUGE-2, and ROUGE-L, respectively.}
\label{tab:in-anatomy-modality}
\end{table*}
To benchmark the performance of different models on the proposed MIMIC-RRS dataset, we conduct experiments within each modality-anatomy pair (\textit{i.e.}, the training and test procedures are performed using only one modality-anatomy pair). We evaluate three types of pre-trained sequence-to-sequence models, namely T5~\cite{raffel2020t5}, BART~\cite{lewis2020bart}, BioBART~\cite{yuan2022biobart}, and their variants.\footnote{We do not evaluate several pre-trained models (e.g., ClinicalBERT~\cite{alsentzer2019publicly}, BioClinicalBERT~\cite{alsentzer2019publicly}, and Clinical-T5~\cite{lu2022clinicalt5}) that specialize in the clinical text since they were trained on the text from MIMIC-III, which overlaps with our dataset. The MIMIC-RRS test set is included in their pre-training data. Thus, we do not adopt them in our experiments to avoid potential data leakage and ensure a fair comparison.} Results are reported in Table~\ref{tab:in-anatomy-modality}.

Several observations can be drawn from these experiments. First, simply adopting pre-trained sequence-to-sequence language models can achieve results comparable to previous state-of-the-art approaches designed for radiology summarization. Indeed, using BART-L as a backbone achieves the best performance, confirming the necessity of exploiting appropriate pre-trained language models.
Secondly, the performances across different model types vary (i.e., BART-L/BART-B, BioBART-L/ BioBART-B). Yet, we notice that the number of training parameters matters; large models report the best results. According to our evaluations, the BART models achieve better results \textit{across all modality-anatomy pairs}.
Surprisingly, it is worth noting that the BioBART models do not achieve better results than BART, although BioBART is pre-trained on a biomedical corpus.
One explanation could be that BioBART models are pre-trained on abstracts from PubMed, which are not within the same domain as radiology reports.

In summary, we note several key findings for future studies: (i) ``\textit{Less is more}'': starting from an appropriate backbone instead of designing complicated modules; (ii) the model size matters; (iii) the pre-training domain matters: knowledge from clinical notes or medical literature does not easily translate to radiology reports.

\subsection{Cross-modality-anatomy} \label{sec:Cross-anatomy-modality}
\begin{figure*}[t]
  \centering\includegraphics[width=1\linewidth]{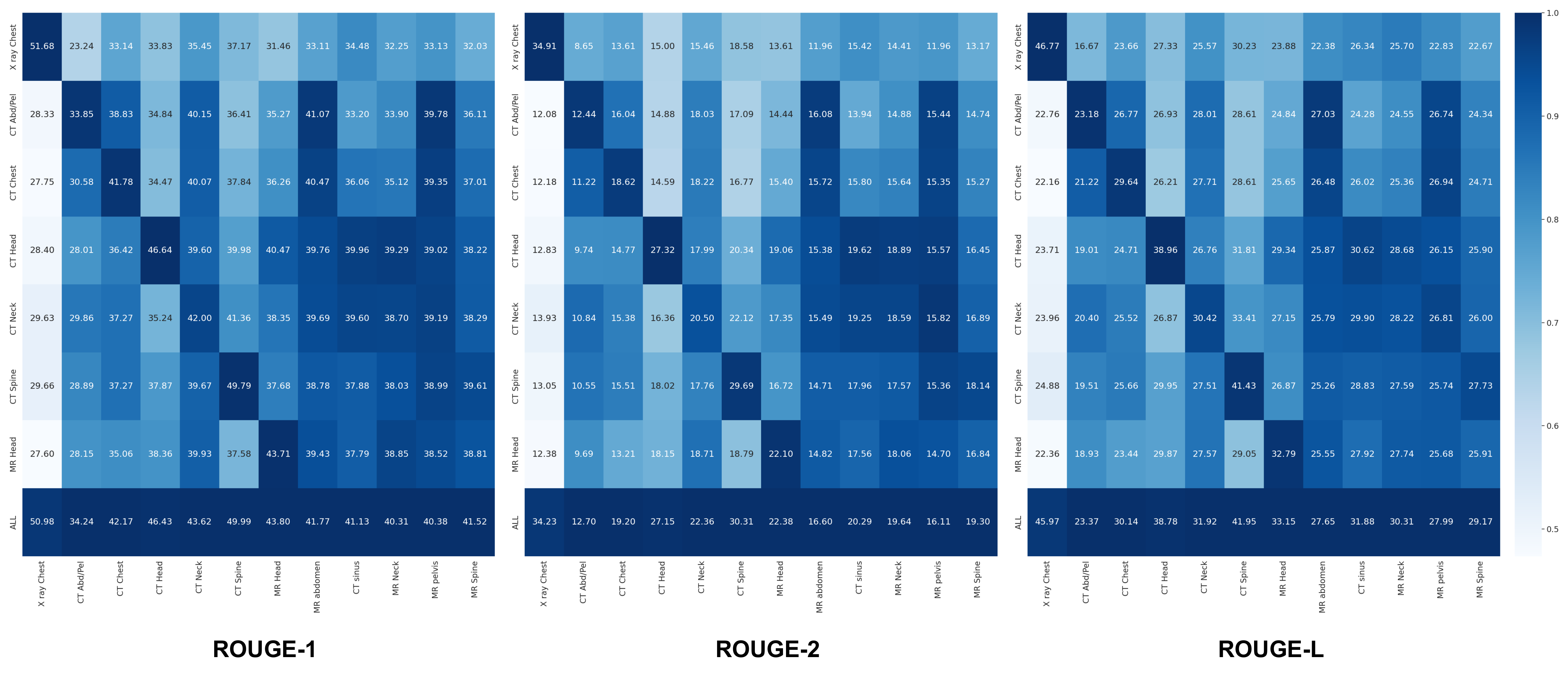}
  \caption{Cross-modality-anatomy results from BART-L are visualized here using heatmaps. Colors from light to dark represent the values from low to high in each column. As discussed in Section~\ref{sec:Cross-anatomy-modality}, the model variant ``ALL'' reports the strongest performances.}
  \label{fig:heatmaps}
\end{figure*}
In this section, we conduct experiments across modality-anatomy pairs (\textit{i.e.}, models are trained on reports from a subset of modality-anatomy pairs and then evaluated on all pairs, including the OOD test sets). We report the cross-modality-anatomy scores in Figure~\ref{fig:heatmaps}. A few interesting observations can be made. First, there are some associations between different anatomies and modalities. For example, the model trained on ``CT Head'' can also achieve promising results on the ``MR Head'' set. Secondly, training the model with all the modality-anatomy pairs (denoted as ALL) achieves the best generalization, obtaining the best results across all modalities and anatomies including the OOD test sets.
We leave further exploration of cross-modality-anatomy associations and zero-shot OOD transfer for future work.

\subsection{Clinical-Efficacy}
\begin{table}[t]
\footnotesize
\centering
\setlength{\tabcolsep}{0.8mm}{
\resizebox{0.99\linewidth}{!}{
\begin{tabular}{@{}lccccc@{}}
\toprule
            & T5-S & BioBART-B & BioBART-L & BART-B        & BART-L        \\ \midrule
MR Head     & 21.5 & 24.8      & 25.3      & 25.0          & \textbf{26.1} \\
CT Spine    & 23.8 & 37.0      & 37.0      & \textbf{38.5} & 38.3          \\
CT Neck     & 21.2 & 23.6      & 23.6      & 24.0          & \textbf{24.9} \\
CT Head     & 31.8 & 34.2      & 34.0      & \textbf{35.2} & 34.7          \\
CT Chest    & 24.0 & 26.0      & 24.3      & \textbf{26.0} & 25.2          \\
CT Abd/Pel  & 12.6 & 15.9      & 15.3      & \textbf{16.1} & 15.9          \\
X-ray Chest & 39.8 & 40.9      & 41.0      & 42.3          & \textbf{43.0} \\ \bottomrule
\end{tabular}}}
\caption{F1-RadGraph scores on MIMIC-RRS test sets across different anatomies and modalities.}
\label{tab:clinical-efficacy}
\end{table}

In addition to evaluating our systems using the ROUGE-1, ROUGE-2, and ROUGE-L metrics~\citep{lin2004rouge}, we use a factual correctness metric to analyze clinical efficacy. Most prior works~\citep{zhang2020optimizing,smit2020combining,hu2022graph} mainly use the {F$_1$CheXbert} metric, an F1-score that evaluates the factual correctness of the generated impressions using 14 chest radiographic observations. Unfortunately, this metric is unsuitable for MIMIC-RRS, which contains reports from other modality-anatomy pairs beyond chest X-rays.

For this reason, instead of using {F$_1$CheXbert}, we propose to use RadGraph~\cite{8ffe9a5} to evaluate the clinical correctness of the generated impressions. RadGraph is a dataset containing board-certified radiologist annotations of radiology reports corresponding to 14,579 entities and 10,889 relations (Appendix~\ref{sec:radgraph}). We used the released pre-trained model to annotate our reports and asked one board-certified radiologist to subjectively validate that the printed entities of the RadGraph model on our data are correct (examples are shown in Table~\ref{table:rad_entities}). After confirming the effectiveness of the model, we follow \citet{delbrouck2022improving} to compute the F1-RadGraph scores. The score evaluates the correctness of the generated named entities in the hypothesis impression compared to the ground-truth impression. We report these results in Table~\ref{tab:clinical-efficacy}. It can be observed that the BART models can achieve the best performance with respect to clinical efficacy. The results are consistent with the ROUGE scores, further confirming the effectiveness of adopting BART as the backbone instead of designing complicated solutions.

\section{Related Work} \label{app:related_work}
In this section, we discuss prior research related to the radiology report summarization task. The first attempt at automatic summarization of radiology findings into natural language impression statements was proposed by \citet{zhang2018learning}. Their contribution was to propose a first baseline on the task, using a bidirectional-LSTM as encoder and decoder. Importantly, they found that about 30\% of the summaries generated from neural models contained factual errors. Subsequently, \citet{zhang2020optimizing} proposed the $\text{F}_1$CheXbert score to evaluate the factual correctness of the generated impression. They also used reinforcement learning to optimize the $\text{F}_1$CheXbert score directly. Finally, both \citet{hu-etal-2021-word} and \citet{hu2022graph} used the Biomedical and Clinical English Model Packages in the Stanza Python NLP Library~\citep{zhang2021biomedical} to extract medical entities. The former study used the entities to construct a Graph Neural Network, which was used as input in their summarization pipeline. In contrast, the latter study used the entities to mask the findings duringcontrastive pre-training.

We believe this paper is an original contribution to the aforementioned line of work. As instigated by \citet{zhang2018learning}, our goal is to release a new summarization corpus and baselines on new modalities and anatomies. We do so by releasing an RRS dataset with data from 11 new modality-anatomy pairs. 
In addition, we extend the work performed by \citet{zhang2020optimizing} by proposing a new metric to evaluates the factual correctness and completeness of the generated impression, namely the RadGraph score. Finally, we improve on the work of \citet{hu-etal-2021-word,hu2022graph} in two ways: (1) we use semantic annotations from a pre-trained model trained using annotations from board-certified radiologists, as opposed to Stanza which leverages unsupervised biomedical and clinical text data; (2) we leverage relation annotations between entities, a feature that was not available in prior work.

\section{Conclusion and Discussion}
In this paper, we highlight and address several weaknesses associated with the radiology report summarization task. First, from a data perspective, we propose a \textit{publicly available} dataset named MIMIC-RRS involving data samples from \textit{twelve} modality-anatomy pairs, with 79,779 samples from MIMIC-III and 128,003 samples from MIMIC-CXR.\\

Second, we conducted more than 40 experiments and over 400 cross-modality-anatomy evaluations to benchmark the performance of different models. We show that instead of designing complicated modules, we can start from an appropriate backbone model such as BART.\\

Finally, we proposed an elegant and simple metric, F1-RadGraph, to evaluate the factual correctness of summaries generated for any modality and anatomy. In the future, we hope that our work broadens the scope of the radiology report summarization task and contributes to the development of reliable RRS models that generalize well to new anatomies and modalities.

\section*{Limitations}
We note two limitations of our paper. First, our work does not extensively evaluate all the available pre-trained models that \textit{could} be suitable for this task, e.g., ELECTRA~\cite{clark2020electra}, BioLinkBERT~\cite{yasunaga2022linkbert}, GatorTron~\cite{yang2022gatortron}, RadBERT~\cite{yan2022radbert}, and PubMedBERT~\cite{gu2021domain}. The aim of this work is not to report the strongest possible score but rather to address weaknesses of existing radiology report summarization studies (in terms of \textit{data} and \textit{evaluation}). Yet, we are confident our proposed solutions report a strong baseline for future work. Second, although F1-RadGraph seems like an appropriate metric to evaluate our new modalities and anatomies (and appears to be consistent with ROUGE scores), it has only been evaluated subjectively and not systematically.

\section*{Acknowledgments}
Maya Varma is supported by graduate fellowship awards from the Department of Defense (NDSEG) and the Knight-Hennessy Scholars program at Stanford University.

\clearpage

\bibliography{custom}
\bibliographystyle{acl_natbib}

\clearpage

\appendix
\section{Details of RadGraph Scores} 
\subsection{The Introduction of RadGraph}\label{sec:radgraph}
\begin{figure}[ht]
  \centering\includegraphics[width=1.0\linewidth]{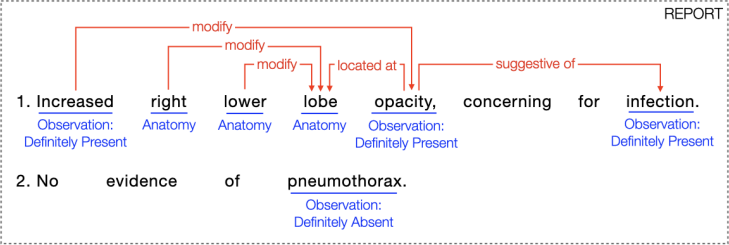}
  \caption{Example of the RadGraph annotations. Figure taken from ~\cite{8ffe9a5}.}
  \label{fig:ddd}
\end{figure}
To design our new evaluation metric, we leverage the RadGraph dataset~\citep{8ffe9a5} containing board-certified radiologist annotations of chest X-ray reports, which correspond to 14,579 entities and 10,889 relations. RadGraph has released a PubMedBERT model~\citep{gu2021domain} pre-trained on these annotations to annotate new reports. An example of annotation can be seen in Figure~\ref{fig:ddd}. Before moving on to the next section, we quickly describe the concept of entities and relations:

\paragraph{Entities} An entity is defined as a continuous span of text that can include one or more adjacent words. Entities in RadGraph center around two concepts: \textit{Anatomy} and \textit{Observation}. Three uncertainty levels exist for \textit{Observation}, leading to four different entities: \textit{Anatomy} (\textit{ANAT-DP}), \textit{Observation: Definitely Present} (\textit{OBS-DP}), \textit{Observation: Uncertain} (\textit{OBS-U}), and \textit{Observation: Definitely Absent} (\textit{OBS-DA}). 

\paragraph{Relations} A relation is defined as a directed edge between two entities. Three levels exist: \textit{Suggestive Of (., .)}, \textit{Located At (., .)}, and \textit{Modify (., .)}. 

\subsection{Metric Computation}\label{sec:score}
Using the RadGraph annotation scheme and pre-trained model, we designed an F-score style reward that measures the factual consistency and completeness of the generated impression (also called hypothesis impression) compared to the reference impression. 

\begin{figure}[!h]
  \centering\includegraphics[width=1.0\linewidth]{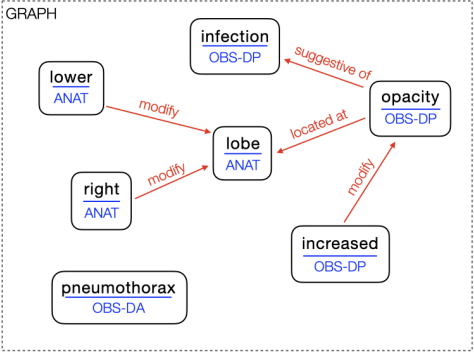}
  \caption{Graph view of the RadGraph annotations for the report in Figure~\ref{fig:ddd}.}
  \label{fig:radgraph}
\end{figure}

To do so, we treat the RadGraph annotations of an impression as a graph $\mathcal{G}(V, E)$ with the set of nodes $V=\{v_1,v_2,\hdots, v_{|V|}\}$ containing the entities and the set of edges $E=\{e_1,e_2,\hdots, e_{|E|}\}$ the relations between pairs of entities. The graph is directed, meaning that the edge $e = (v_1, v_2) \ne (v_2, v_1)$. An example is depicted in Figure~\ref{fig:radgraph}. Each node or edge of the graph also has a label, which we denote as $v_{i_{L}}$ for an entity $i$ (for example ``OBS-DP" or ``ANAT") and $e_{ij_{L}}$ for a relation $e = (v_i, v_j)$ (such as ``modified" or ``located at").

To design our RadGraph score, we focus on the nodes $V$ and whether or not a node has a relation in $E$. For a hypothesis impression $y$, we create a new set of triplets ${T}_y = \{{(v_i, v_{i_{L}}, \mathcal{R})}\}_{i=1:{|V|}}$. The value $\mathcal{R}$ is $1$ if $(v_i, v_j)_{j=1:{|E|}, i\ne j} \in E$, $0$ otherwise. In other words, a triplet contains an entity, the entity label, and whether or not this entity has a relation. We proceed to construct the same set for the reference report $\hat{y}$ and denote this set ${T}_{\hat{y}}$. 

Finally, our score is defined as the harmonic mean of precision and recall between the hypothesis set ${T}_{{y}}$ and the reference set $T_{\hat{y}}$, giving a value between $0$ and $100$. As an illustration, the set $V$, $E$ and $T$ of the graph $\mathcal{G}$ in Figure~\ref{fig:radgraph} are shown as follows:

$V=\{$mild, fluid, overload, overt, pulmonary, edema$\}$

$E=\{$(mild,overload), (overload, fluid), (edema, pulmonary)$\}$

$T=\{$(mild, obs-dp, 1), (fluid, obs-dp, 0), (overload, obs-dp, 1), (overt, obs-da, 0), (pulmonary, anat-dp, 0), (edema, obs-da, 1)$\}$

\section{Code and Data Release} \label{app:code}
Our research has been carried out using the ViLMedic library~\citep{delbrouck-etal-2022-vilmedic}. Our code is available at \url{https://github.com/jbdel/vilmedic}. This link is anonymized and complies with the double-blind review process. More specifically, we release the code of the RadGraph score as well as the training of our baseline. We also release the script to download, pre-process, and split the radiology reports of the MIMIC-III database as per our experiments.
To download the MIMIC-III database, researchers are required to formally request access via a process documented on the MIMIC website. There are two key steps that must be completed before access is granted:
(i) the researcher must complete a recognized course in protecting human research participants, including Health Insurance Portability and Accountability Act (HIPAA) requirements.
(ii) the researcher must sign a data use agreement, which outlines appropriate data usage and security standards, and forbids efforts to identify individual patients.

\section{More Results}
\begin{figure*}[t]
  \centering\includegraphics[width=1\linewidth]{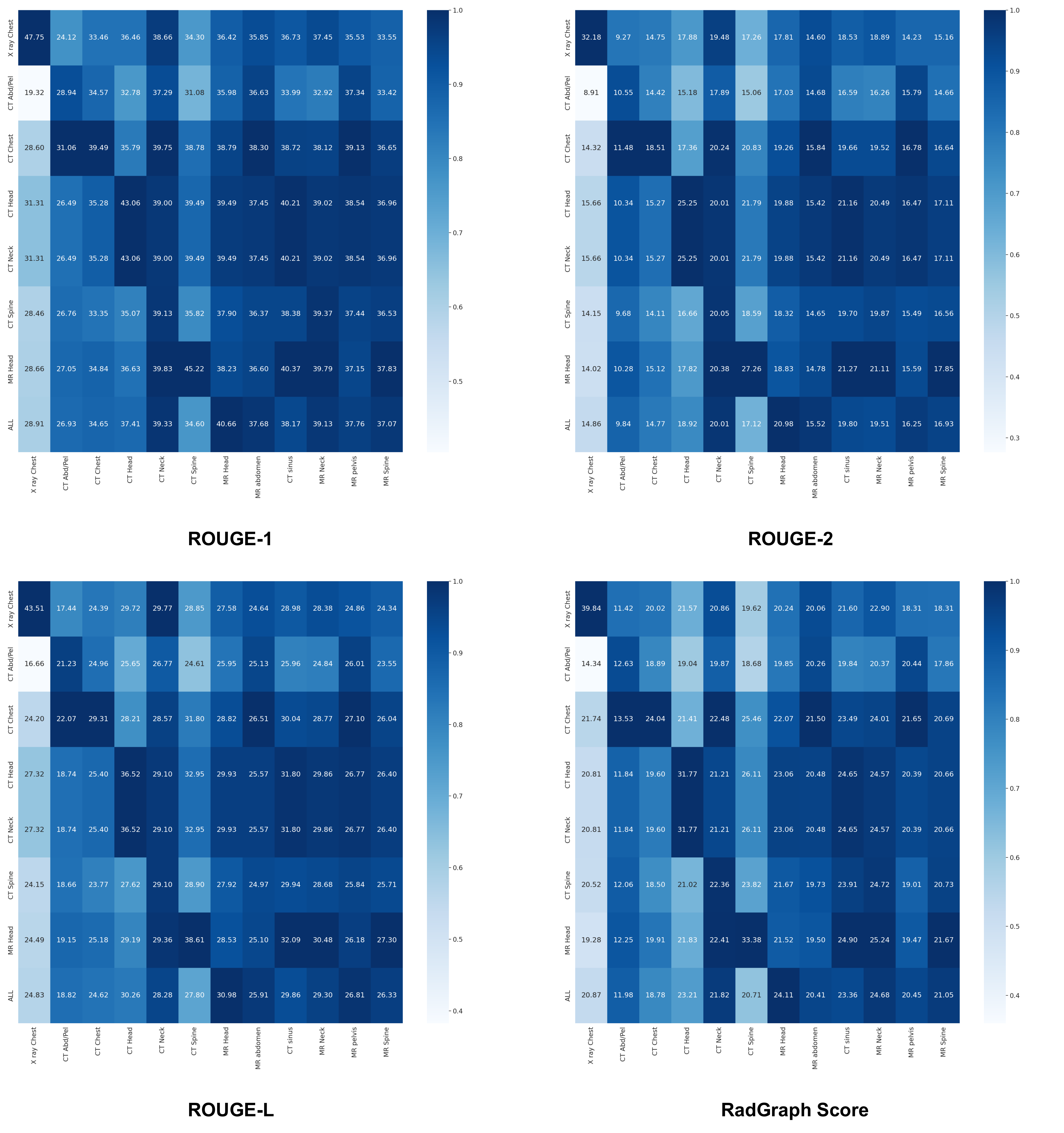}
  \caption{Cross-modality-anatomy results from T5-S are visualized here using heatmpas. Colors from light to dark represent the values from low to high in each column.}
  \label{fig:heatmaps-t5}
\end{figure*}

\begin{figure*}[t]
  \centering\includegraphics[width=1\linewidth]{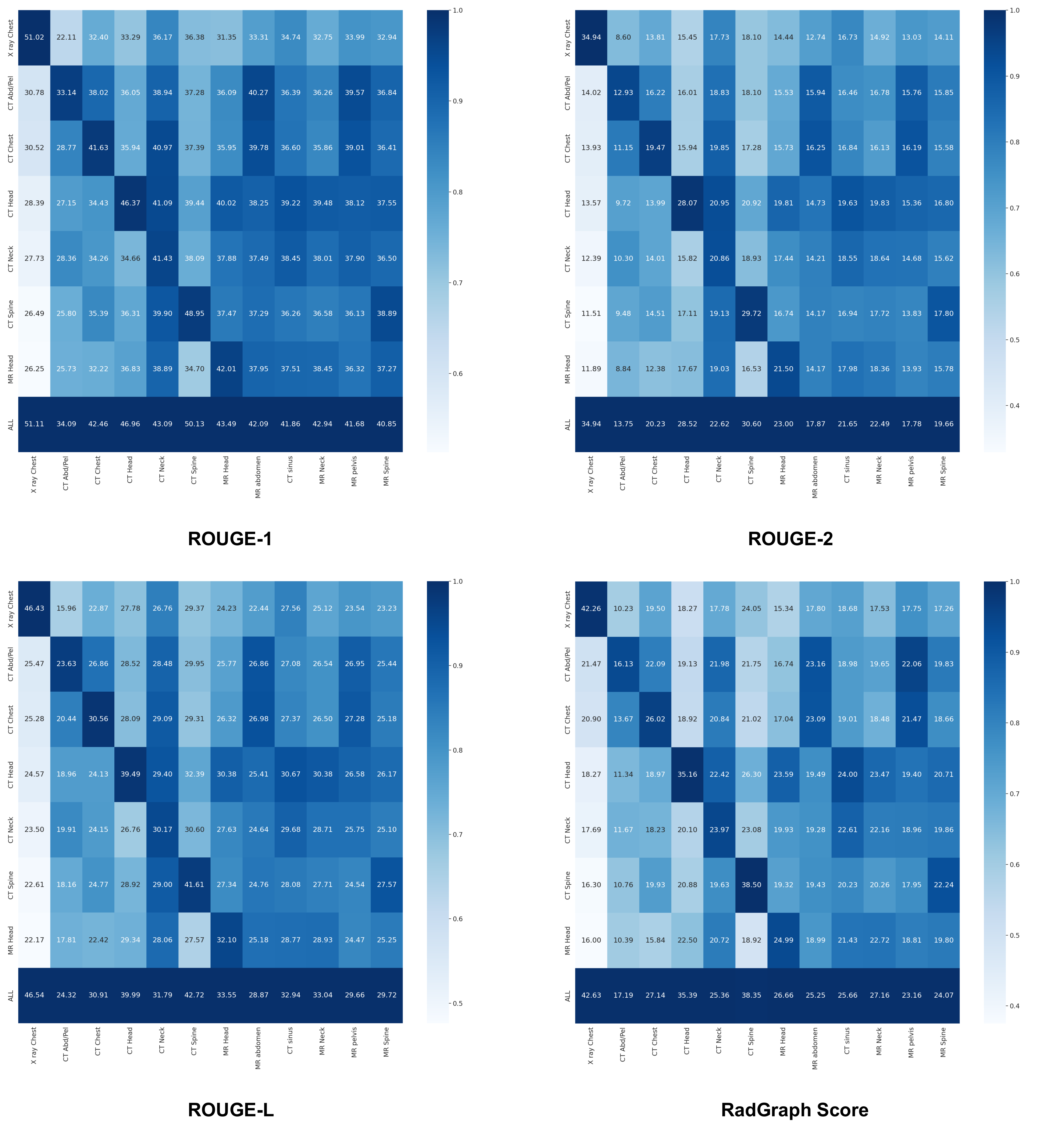}
  \caption{Cross-modality-anatomy results from BART-B are visualized here using heatmaps. Colors from light to dark represent the values from low to high in each column.}
  \label{fig:heatmaps-bart-base}
\end{figure*}

\begin{figure*}[t]
  \centering\includegraphics[width=1\linewidth]{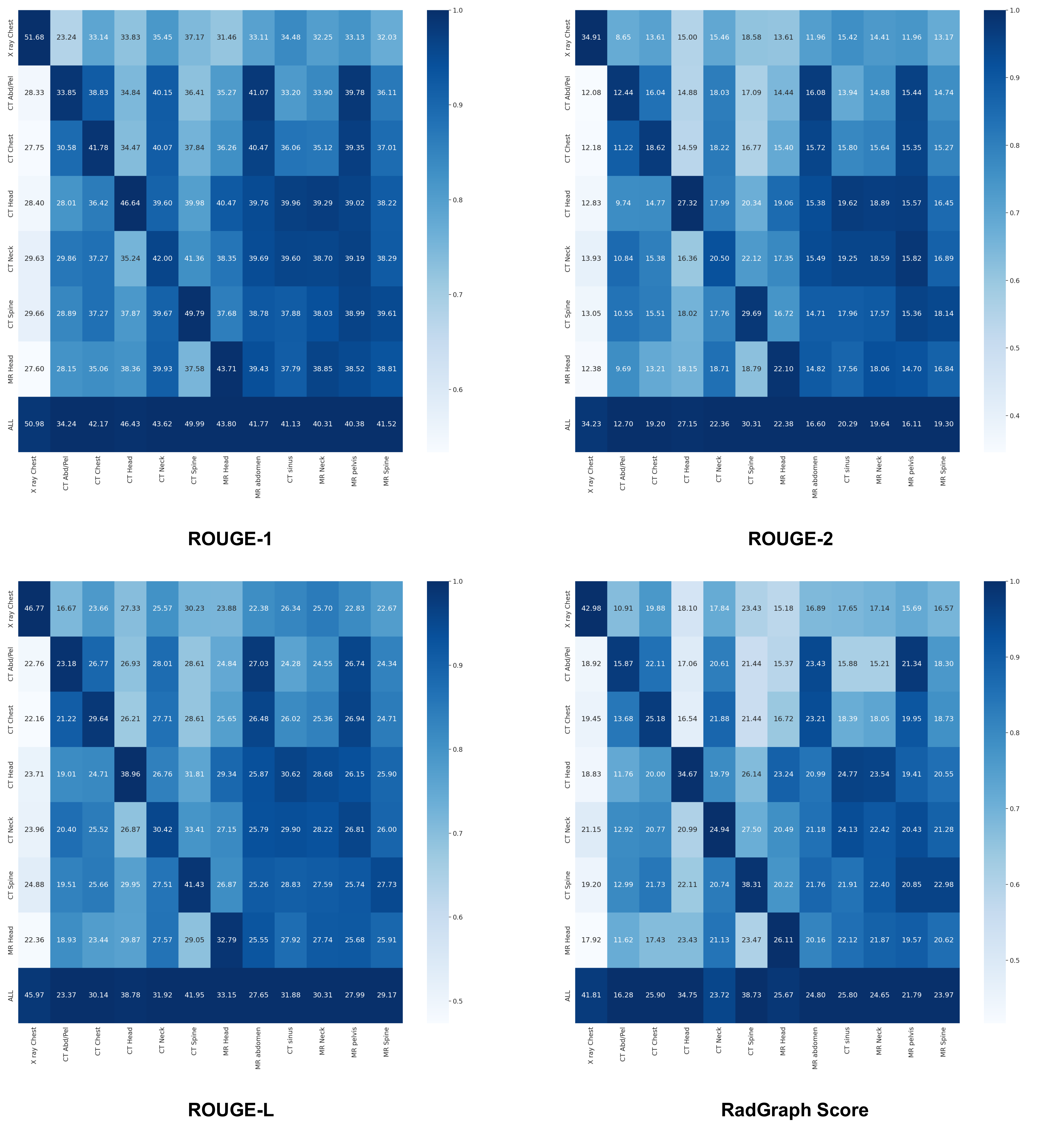}
  \caption{Cross-modality-anatomy results from BART-L are visualized here using heatmaps. Colors from light to dark represent the values from low to high in each column.}
  \label{fig:heatmaps-bart-large}
\end{figure*}

\begin{figure*}[t]
  \centering\includegraphics[width=1\linewidth]{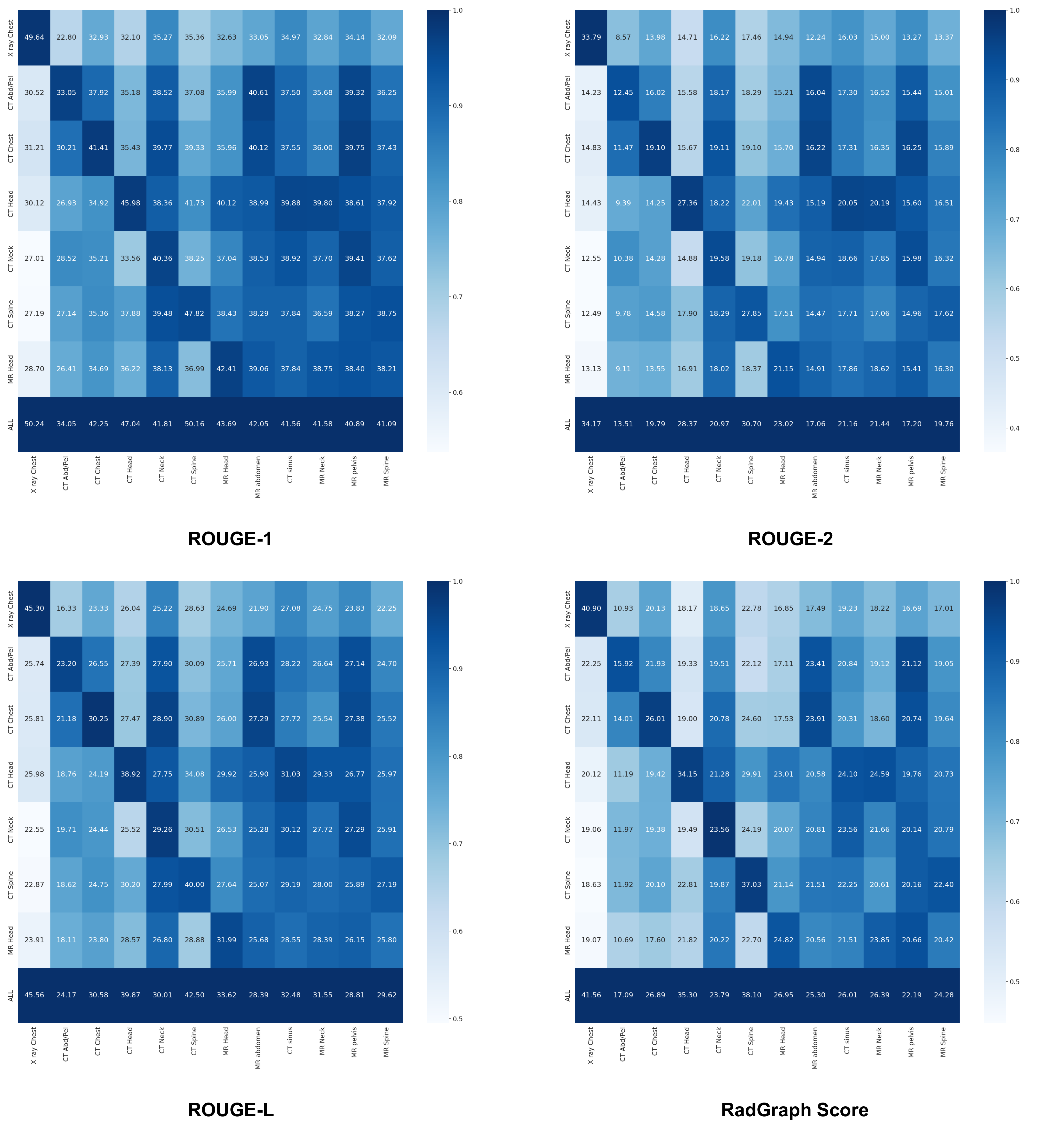}
  \caption{Cross-modality-anatomy results from BioBART-B are visualized here using heatmaps. Colors from light to dark represent the values from low to high in each column.}
  \label{fig:heatmaps-biobart-base}
\end{figure*}

\begin{figure*}[t]
  \centering\includegraphics[width=1\linewidth]{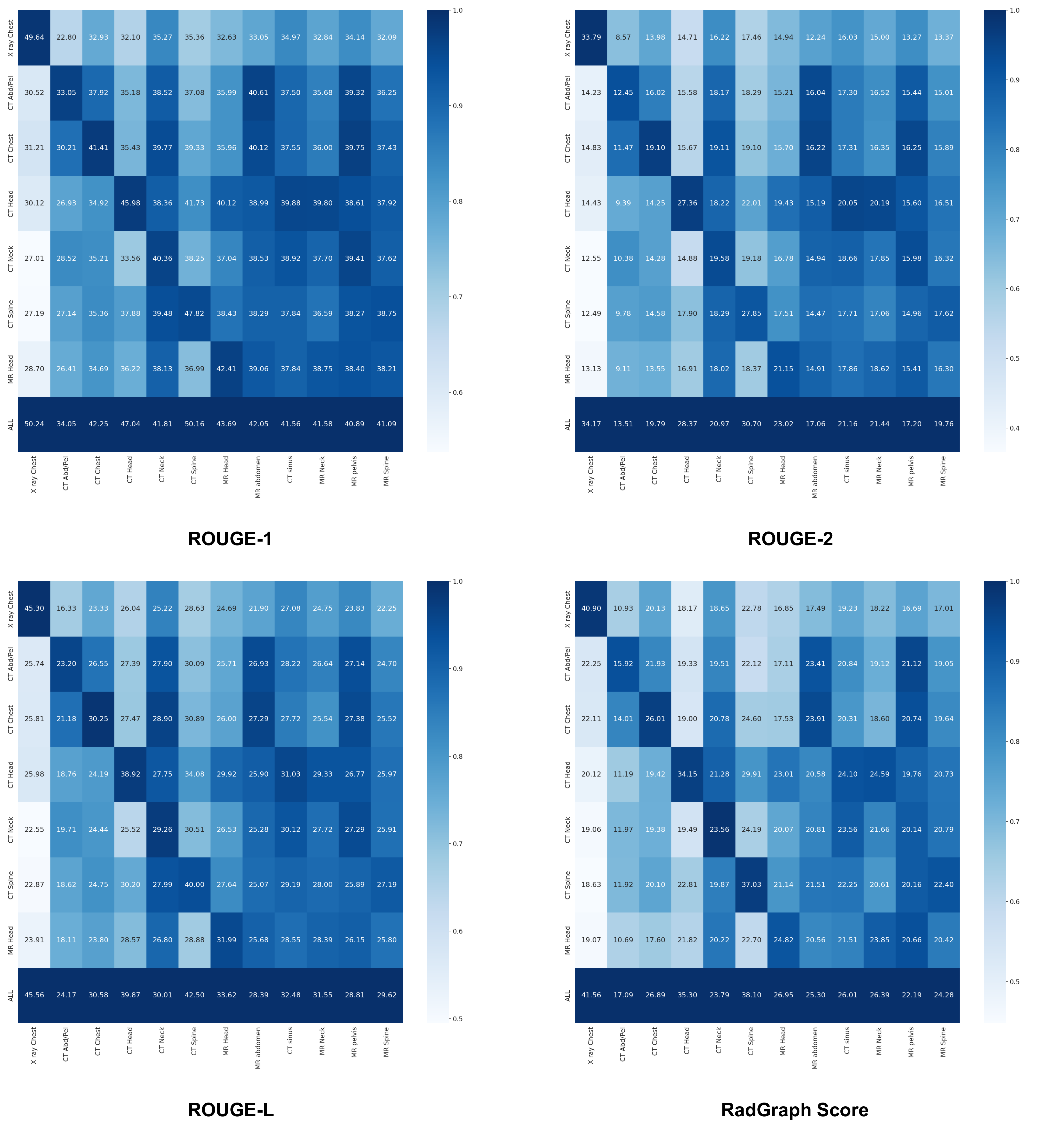}
  \caption{Cross-modality-anatomy results from BioBART-L are visualized here using heatmaps. Colors from light to dark represent the values from low to high in each column.}
  \label{fig:heatmaps-biobart-large}
\end{figure*}

We present the results (including four metrics, \textit{i.e.}, ROUGE-1, ROUGE-2, ROUGE-L, and RadGraph scores) of all the experiments on Figure~\ref{fig:heatmaps-t5}-\ref{fig:heatmaps-biobart-large} for further research in this field. We also show the output of RadGraph (for entities) on a few samples of our new dataset in Table~\ref{table:rad_entities}.

 \begin{table*}[!t]
\resizebox{\textwidth}{!}{
\begin{tabular}{Sc Sc Sc Sc}
\hline
CT Spine & CT Sinus & MR Neck & MR Head\\ \hline
\cincludegraphics[width=0.20\linewidth]{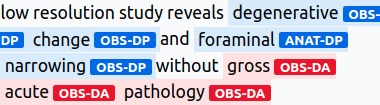} & \cincludegraphics[width=0.25\linewidth]{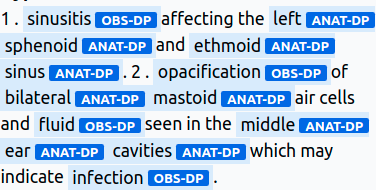} &  \cincludegraphics[width=0.25\linewidth]{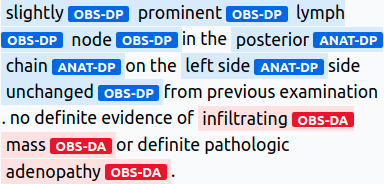} &   
\cincludegraphics[width=0.25\linewidth]{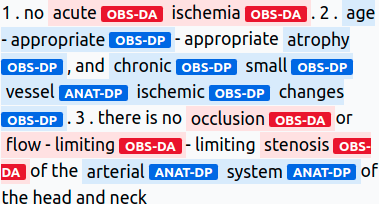}  \\ \\ \hline
\end{tabular}
}
\caption{Examples of entites detected by RadGraph (used in the RG$_{\text{ER}}$ metric) on out-of-domain anatomy/modality radiology reports. Relations are omitted for clarity.}
\label{table:rad_entities}
\end{table*}


\section{Ethics Statement}
The MIMIC-CXR and MIMIC-III datasets are de-identified to satisfy the US Health Insurance Portability and Accountability Act of 1996 (HIPAA) Safe Harbor requirements. Protected health information (PHI) has been removed. Therefore, the ethical approval statement and the need for informed consent were waived for the studies on this database, which was approved by the Massachusetts Institute of Technology (Cambridge, MA) and Beth Israel Deaconess Medical Center (Boston, MA). This research was conducted in accordance with the Declaration of Helsinki, describing the ethical principles of medical research involving human subjects.

\end{document}